\documentclass{article}

\usepackage{PRIMEarxiv}

\usepackage[utf8]{inputenc} 
\usepackage[T1]{fontenc}    
\usepackage{hyperref}       
\usepackage{url}            
\usepackage{booktabs}       
\usepackage{amsfonts}       
\usepackage{nicefrac}       
\usepackage{microtype}      
\usepackage{lipsum}
\usepackage{fancyhdr}       
\usepackage{graphicx}       
\usepackage{xcolor,colortbl}
\usepackage{multirow}
\usepackage{makecell}
\usepackage{xspace}
\usepackage{authblk}
\usepackage{}
\usepackage{ifthen}  
\usepackage{pgffor} 
\usepackage[symbol]{footmisc}
\graphicspath{{media/}}     

\makeatletter
\DeclareRobustCommand\onedot{\futurelet\@let@token\@onedot}
\def\@onedot{\ifx\@let@token.\else.\null\fi\xspace}

\def\eg{\emph{e.g}\onedot}

\def\etc{\emph{etc}\onedot} 
 
\def\etal{\emph{et al}\onedot}
\makeatother

\title{
\textbf{InfiMM-Eval}: Complex Open-ended Reasoning Evaluation for Multi-modal Large Language Models
}

\author[1]{Xiaotian Han}
\author[1]{Quanzeng You}
\author[1]{Yongfei Liu}
\author[1]{Wentao Chen}
\author[1, $\ddagger$]{Huangjie Zheng}
\author[1]{Khalil Mrini}
\author[1]{Xudong Lin}
\author[1, $\ddagger$]{Yiqi Wang}
\author[1]{Bohan Zhai}
\author[1]{Jianbo Yuan}
\author[1]{Heng Wang}
\author[1]{Hongxia Yang}

\affil[1]{ByteDance Inc., \{xiaotian.han, quanzeng.you, hx.yang\}@bytedance.com}

\begin{document}
\maketitle
\begin{abstract}
Multi-modal Large Language Models (MLLMs) are increasingly prominent in the field of artificial intelligence. 
These models not only excel in traditional vision-language tasks but also demonstrate impressive performance in contemporary multi-modal benchmarks.
Although many of these benchmarks attempt to holistically evaluate MLLMs, they typically concentrate on basic reasoning tasks, often yielding only simple yes/no or multi-choice responses. 
These methods naturally lead to confusion and difficulties in conclusively determining the reasoning capabilities of MLLMs. 
To mitigate this issue, we manually curate a benchmark dataset specifically designed for MLLMs, with a focus on complex reasoning tasks.
Our benchmark comprises three key reasoning categories: deductive, abductive, and analogical reasoning.
The queries in our dataset are intentionally constructed to engage the reasoning capabilities of MLLMs in the process of generating answers.
For a fair comparison across various MLLMs, we incorporate intermediate reasoning steps into our evaluation criteria. 
In instances where an MLLM is unable to produce a definitive answer, its reasoning ability is evaluated by requesting intermediate reasoning steps. 
If these steps align with our manual annotations, appropriate scores are assigned. 
This evaluation scheme resembles methods commonly used in human assessments, such as exams or assignments, and represents what we consider a more effective assessment technique compared with existing benchmarks.
We evaluate a selection of representative MLLMs using this rigorously developed open-ended multi-step elaborate reasoning benchmark, designed to challenge and accurately measure their reasoning capabilities. The code and data will be released at \textcolor{red}{\href{https://infimm.github.io/InfiMM-Eval/}{https://infimm.github.io/InfiMM-Eval/}}.
\end{abstract}


\footnotetext[3]{Work done during internship at ByteDance.}

\section{Introduction}
\label{sec:intro}
The field of natural language processing (NLP) has been profoundly transformed by the emergence of large language models (LLMs)~\cite{vaswani2017attention,min2023recent}. 
Exhibiting exceptional proficiency in a wide range of NLP tasks~\cite{devlin2018bert,radford2019language}, LLMs have led to the development of Multi-modal Large Language Models (MLLMs), which combine language processing with other modalities, primarily visual modality, enhancing content understanding and generation across domains~\cite{alayrac2022flamingo, rombach2022high,driess2023palm,ghosal2023text}.

\begin{figure}
    \centering
    \includegraphics[width=0.9\textwidth]{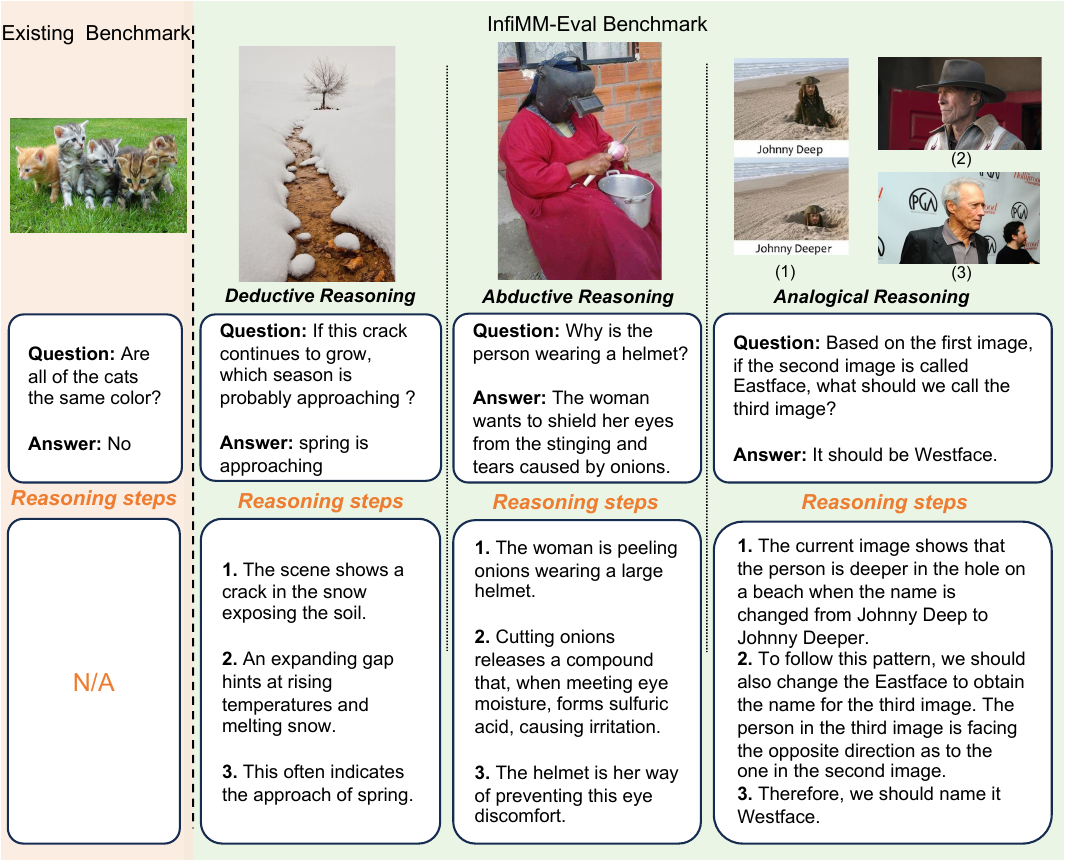}
    \caption{Comparison between existing MLLM benchmarks and our InfiMM-Eval. \textbf{Left}: Existing benchmarks usually involve basic reasoning tasks and simple responses. \textbf{Right}: InfiMM-Eval benchmark consists of deductive, abductive, and analogical reasoning categories. Each sample includes one or more images, one question, one answer, and the reasoning steps to deduce the answer.}
    \label{fig:teaser}
    \vspace{-10pt}
\end{figure}

Leading in-house models like Flamingo~\cite{alayrac2022flamingo}, Palm-e~\cite{driess2023palm}, RT-2~\cite{brohan2023rt}, and GPT-4V(ision)~\cite{gpt4systemcard} have exemplified the extensive applicability and promising potential of MLLMs. 
The open-source community has also contributed significantly to the field through the development of innovative architectures and the creation of curated instruction fine-tunning datasets, including MiniGPT-4~\cite{zhu2023minigpt}, LLaVA~\cite{liu2023llava}, IDEFICS~\cite{laurenccon2023obelisc}, \etc. 
Each model provides distinct insights, exploring a variety of aspects and potential applications of multi-modal interactions. 

Several studies have explored LLMs, highlighting their potential~\cite{chang2023survey,guo2023evaluating}.
However, as noted in~\cite{wei2022emergent}, their performance, especially in reasoning tasks, often escalates unpredictably.
Reasoning, a key component for human-level intelligence ~\cite{mccarthy2007here,darwiche2018human}, is challenging to evaluate, leading to the development of specific benchmarks such as ARB~\cite{sawada2023arb}, ARC~\cite{Clark2018ThinkYH}, and GSM8k~\cite{cobbe2021training}. 
For MLLMs, visual understanding extends beyond mere perception~\cite{zellers2019recognition}, the need for specialized reasoning benchmarks is even more critical.

Recent advancements in the Multimodal Large Language Models (MLLMs) research field have led to the establishment of comprehensive evaluation benchmarks such as MME~\cite{fu2023mme}, MMBench~\cite{liu2023mmbench}, SeedBench~\cite{li2023seed}, and MathVista~\cite{lu2023mathvista}. 
While reasoning ability is a crucial factor assessed in these benchmarks, there is variation in how they categorize the reasoning capabilities of MLLMs, which could lead to potential confusion and challenges in gaining clear insights.
In addition, existing benchmarks, predominantly centered on visual commonsense reasoning such as VCR~\cite{zellers2019recognition}, or those that transform tasks into a multiple-choice format to streamline evaluation, may not sufficiently challenge advanced models such as GPT-4V.
This suggests a need for more stringent and comprehensive benchmark to thoroughly evaluate the reasoning capabilities of Multimodal Large Language Models.

To address the issues identified above, we introduce the \textbf{InfiMM-Eval} benchmark. 
This benchmark is designed to evaluate open-ended complex visual reasoning problems.
Drawing on the work of \cite{Conner2014IdentifyingKO} in the field of logical reasoning, we categorize samples into three reasoning paradigms: deductive, abductive, and analogical reasoning. 
\figureautorefname~\ref{fig:teaser} presents examples from each of these reasoning categories.
Such categorization encompasses a broad range of practical applications in reasoning and thus offers comprehensive insights into the reasoning capabilities of  MLLMs.
Our benchmark additionally includes detailed sequential steps employed in the reasoning process to answer each question.
These reasoning steps are pivotal in assessing the reasoning capabilities of models, particularly in complex real-world scenarios. 
To the best of our knowledge, InfiMM-Eval represents the first multi-modal, open-ended QA benchmark that incorporates such detailed reasoning steps.

Moreover, the inclusion of reasoning steps facilitates the creation of a more sophisticated evaluation protocol. 
Following rubric grading format, we design our assessment protocol as: the response receives full marks for a directly correct answer, and partial scores are allocated based on the relevance and logic of its intermediate reasoning steps.
This method not only underscores the model's proficiency in generating accurate answers but also provides a thorough analysis of its decision-making process, thereby elucidating its reasoning pathways.
We employ an LLM-based evaluator to implement this evaluation protocol for open-ended responses that include reasoning steps.

\noindent Our contributions can be summarized as follows:
\begin{itemize}
    \item We present InfiMM-Eval, a manually curated high-quality benchmark with complex reasoning questions designed specifically for evaluating MLLMs.
    \item We propose to evaluate open-ended MLLM reasoning response by combining intermediate reasoning steps and final answers for intricate scoring.
    \item We perform ablation studies on representative MLLMs to evaluate their reasoning capabilities using our InfiMM-Eval benchmark.
\end{itemize}
\section{Related work}
\subsection{Multi-modal LLMs}
The evolution of LLMs has inspired research on integrating visual signal into LLMs. 
For example, Flamingo~\cite{alayrac2022flamingo} integrates the Perceiver~\cite{jaegle2021perceiver} Resampler and gated attention modules onto LLMs, bridging visual encoders and LLMs, thereby proving highly effective in in-context learning for vision-language tasks.
Other giant models like Palm-e~\cite{driess2023palm}, RT-2~\cite{brohan2023rt}, and GPT-4V(ision)~\cite{gpt4systemcard} have also underscored the expansive applicability and potential of MLLMs.

Various smaller-sized MLLMs have emerged recently.
Mini-GPT4~\cite{zhu2023minigpt4} utilizes the instruction-tuned Vicuna~\cite{vicuna2023},  and fine-tunes a linear layer to align vision and language representations.
LLaMA-Adapter~\cite{zhang2023llamaadapter} introduces a lightweight adapter to enable the adaptability of LLaMA to visual inputs. 
BLIP-2~\cite{li2023blip2} incorporates the Q-Former, adding a crucial alignment stage to connect the frozen LLM with the visual modality, notably excelling in Visual Question Answering (VQA) tasks.
InstructBLIP~\cite{dai2023instructblip} focuses on fine-tuning the Q-Former using diverse instruction tuning datasets, enhancing its performance in visual scene comprehension and visual dialogues. 
In contrast, Otter~\cite{li2023otter}, refines the OpenFlamingo~\cite{awadalla2023openflamingo} for improved instruction-following capabilities and more effective usage of in-context samples.
Multimodal-CoT~\cite{zhang2023multimodal} integrates chain-of-thought~\cite{kojima2022large,wei2022chain} into the multimodal domain, showcasing robust results on the ScienceQA benchmark. 
MMICL~\cite{zhao2023mmicl} tackles the challenges posed by multi-modal inputs with multiple images, targeting intricate multi-modal prompts and detailed text-to-image references. 
LLaVA~\cite{liu2023llava} employs a simple linear connector and fine-tunes the entire LLM to boost performance. Its enhanced version, LLaVA-1.5~\cite{liu2023improved}, integrates large-scale instruction tuning and high-resolution images, achieving superior results across various benchmarks.

\subsection{MLLM evaluation benchmarks}
Different vision-language benchmarks have been introduced to evaluate the specific reasoning capabilities of MLLMs. 
For instance, Winoground~\cite{thrush2022winoground} assesses the visual-linguistic compositional reasoning, RAVEN~\cite{zhang2019raven} focuses on relational and analogical reasoning, OK-VQA~\cite{marino2019okvqa} examines reasoning with external knowledge, and VCR~\cite{zellers2019vcr} evaluates visual commonsense reasoning related to people in video frames. 
Other benchmarks, such as TextVQA~\cite{singh2019towards}, FigureQA~\cite{kahou2018figureqa}, and ScienceQA~\cite{saikh2022scienceqa}, have also made significant contributions by addressing reasoning within diverse contexts.
MathVista~\cite{lu2023mathvista} provides a consolidated assessment of mathematical reasoning capabilities.

In addition to the above-mentioned reasoning-specific benchmarks, comprehensive benchmarks have been proposed, which also include assessments of various reasoning capabilities. 
For instance, MME~\cite{fu2023mme} evaluates reasoning capabilities of commonsense reasoning, numeric calculation, text translation, and code understanding.
MMBench~\cite{liu2023mmbench} assesses logical, attribute, and relation reasoning, while SEED-Bench~\cite{li2023seedbench} contains visual reasoning, action prediction, and procedure understanding. 
All above benchmarks use multiple-choice question format to simplify the evaluation process. This leads to unnatural questioning and models may obtain hints from choices. 
On the other hand, scoring by final answer correctness only underestimates the importance of reasoning process, which is not enough to understand the models' reasoning capability.

Thus, open-ended benchmarks are needed to better align with the generative nature of recent MLLMs. However, traditional metrics, like CIDEr~\cite{vedantam2015cider}, SPICE~\cite{anderson2016spice}, \etc are not suitable for open-ended QA evaluation. 
Human evaluations are prohibitively costly.
Luckily, Chiang~\etal~\cite{chiang2023large} suggest LLMs can be an alternative to human evaluators. 
Recent open-ended QA benchmarks for MLLMs, such as TouchStone~\cite{bai2023touchstone}, VisIT-Bench~\cite{bitton2023visitbench}, and MM-Vet~\cite{yu2023mmvet}, also employ LLM-based evaluators.
This further demonstrates the reliability of LLM-based evaluators in such context.

\subsection{Reasoning in MLLMs}
Human reasoning, essential for intelligence, involves analyzing information to derive logical insights~\cite{yu2023nature, huang2022towards, walton1990reasoning}.
LLMs have demonstrated substantial reasoning abilities in NLP tasks, as evidenced in recent studies ~\cite{kojima2022large,huang2022towards,wei2022emergent,yao2022react,webb2023emergent}.
Similar capabilities are observed in~\cite{driess2023palm,gpt4systemcard}.
However, MLLMs research field lacks a systematic and unified framework for categorizing reasoning capability. 
Current benchmarks fragment reasoning into numerous task-specific categories, \eg commonsense reasoning, math reasoning, code understanding, procedure understanding \etc. 
Such categorization may potentially obscure a holistic understanding of the reasoning capacities of MLLMs.
Our study advocates for a directional classification of reasoning in MLLMs, anchored in established logical principles~\cite{bronkhorst2020logical,dowden2018logical}, focusing on deductive, abductive, and analogical reasoning, essential in human cognition.

\noindent\textbf{Deductive reasoning} derives new conclusions from established premises~\cite{johnson1999deductive}, ensuring that the steps of inference align with established logical rules.
To illustrate, consider the deductive example presented on the right of \figureautorefname~\ref{fig:teaser}: the premises include observations as ``snow is presented in image'', ``soil is revealed after snow melting, looks like crack'', and ``crack is expanding''. From these premises, the deductive conclusion drawn from premises is ``current season is winter, after winter it will be spring''.
Deductive reasoning capability is vital for MLLMs in various domains. 
This encompasses automatic fact-checking of multi-modal information and multi-modal legal reasoning for interpreting legal documents, among other applications.

\noindent\textbf{Abductive reasoning} determines the most plausible explanation, grounded in common sense for a specific set of observations \cite{douven2011abduction}.
This form of reasoning is often viewed as the converse of deductive reasoning.
in the abductive scenario illustrated in \figureautorefname~\ref{fig:teaser}, the observation is ``a person is cutting an onion while wearing a helmet''. 
Given the commonsense knowledge that ``Onions can release compounds causing eyes irritation'', the most plausible explanation for the question is ``eye protection''.
The capability of abductive reasoning extends to causal inference in complex systems.
It can be applied, but is not limited to, inferring public sentiment from economic data and news, or predicting trends from text, images, and videos.

\noindent\textbf{Analogical reasoning} facilitates the transfer of knowledge from known instances to analogous situations \cite{goswami1991analogical}. 
In the example illustrated in \figureautorefname~\ref{fig:teaser}, the first image demonstrates a proposition that the naming convention is a play on words involving depth. The second and third images should adhere to a similar pattern. 
Specifically, while the individual in the second image is facing east, the person in the third image faces west, suggesting that his name should logically be ``Westface''.
The capability for analogical reasoning is pivotal in comparative analysis, which constitutes a fundamental aspect of in-context learning.

In this work, we introduce InfiMM-Eval, a novel open-ended QA benchmark, dedicated to assessing the reasoning capabilities of MLLMs, with systematically designed and categorized reasoning questions. 

\section{InfiMM-Eval benchmark}
\subsection{Data collection}
Compared with the extensive, automatically collected MLLM reasoning datasets as discussed in prior studies~\cite{li2023otter,liu2023llava, zhao2023svit}, our InfiMM-Eval initiative is dedicated to the manual creation of a high-quality evaluation benchmark. 
This benchmark is particularly designed to evaluate the multi-step reasoning abilities increasingly evident in contemporary MLLMs.
It specifically emphasizes deductive, abductive, and analogical reasoning, which are fundamental to routine human cognitive processes.

In alignment with this principle, the process of collecting data for our evaluation benchmark can be broadly categorized into the following steps:

\noindent\textbf{Question and answer collection.} Our methodology involved engaging eight annotators, each tasked with sourcing a wide range of images from varied scenarios.
These images were sourced from a variety of platforms, including online platforms and existing public dataset, notably adopting 25 samples from MM-Vet~\cite{yu2023mmvet}.
The primary objective for these annotators was to create a comprehensive set of questions and answers. 
It was imperative that these questions were crafted to rigorously test the multi-step logical reasoning capabilities of MLLMs. 
To ensure the complexity of the task, the questions were designed to be intricate enough to preclude the possibility of immediate answers based purely on visual observation.

To ensure the robustness of this study, specific guidelines were established for the formulation of questions.
Although the answers format were permitted a degree of openness, the questions themselves were required to have a single logic path. 
This means that despite the potential openness in responses, the line of reasoning to arrive at these answers should be fairly consistent among different individuals. 
For example, overly subjective questions like ``What is your feeling when you see this image?'' were excluded.
These types of questions do not align with the standard of robustly eliciting a logical reasoning pathway.

Additionally, each sample was meticulously categorized into one of three distinct reasoning types: deductive, abductive or analogical. 
This classification not only aids in organizing the dataset but also ensures a comprehensive assessment of various reasoning skills.
\begin{figure}
    \centering
    \includegraphics[width=0.8\textwidth]{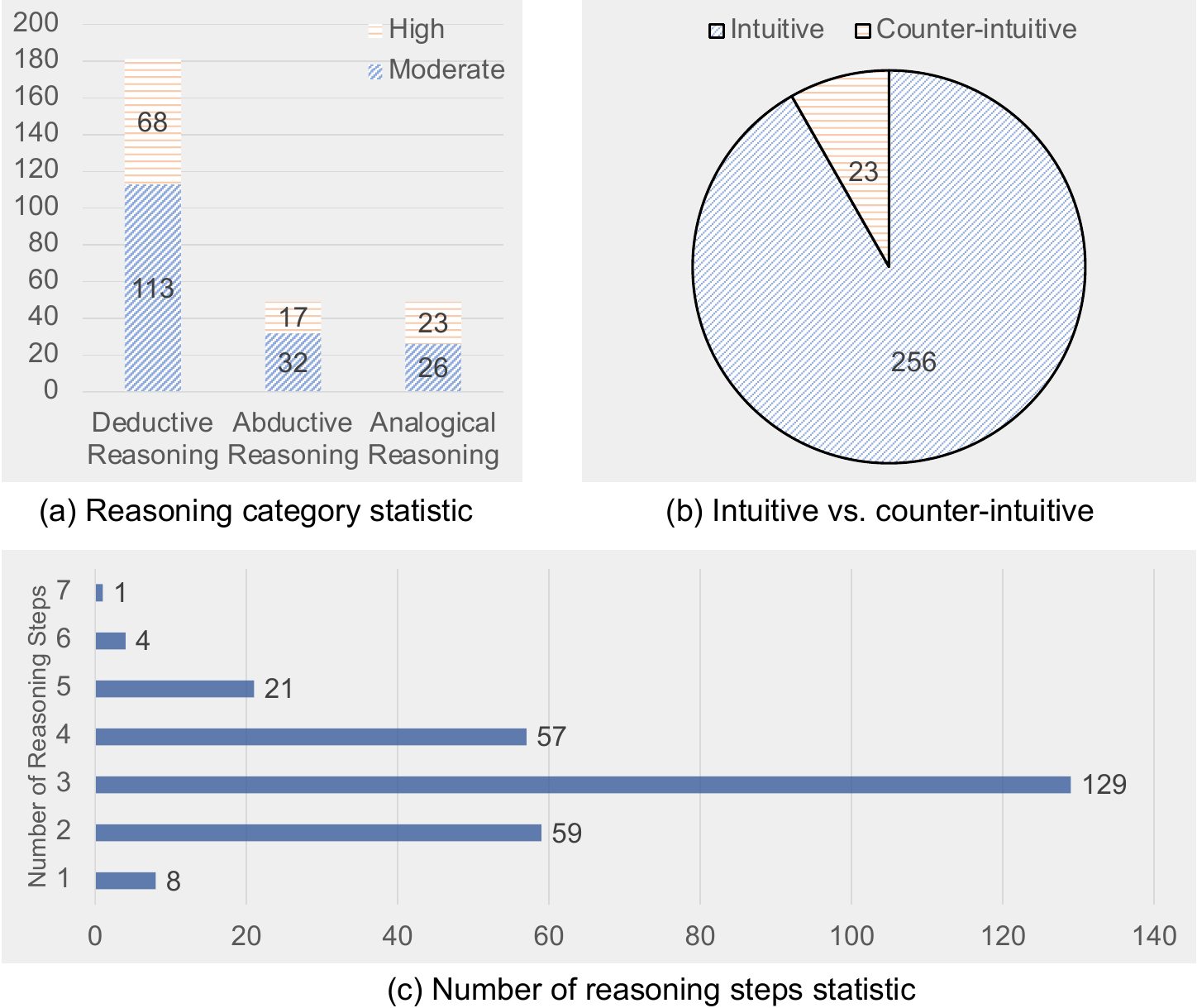}
    \caption{InfiMM-Eval benchmark statistics: (a) indicates distribution of reasoning categories and their respective reasoning complexity; (b) represents the statistic of counter-intuitive versus intuitive reasoning questions; and (c) shows the breakdown of the number of reasoning steps per question.
    }
    \label{fig:data_cate_stats}
\end{figure}

\noindent\textbf{Quality control.} To guarantee the exceptional quality of our benchmark, we implemented a thorough cross-validation protocol. 
Each sample underwent validation by two independent annotators. 
Their evaluation is based on a comprehensive set of standards, which includes:
\begin{itemize}
    \item \textbf{Appropriateness check:} Each image and question is examined for inappropriate or offensive content, ensuring fairness, diversity, and suitability for a diverse audience.
    \item \textbf{Consistency analysis:} The relationship between the question, answer, and reasoning steps are carefully evaluated to ensure they are logically aligned and coherent.
    \item \textbf{Image relevance:} This criterion assesses whether the image is essential for answering the question, thereby filtering samples where questions could be answered without the visual aid.
    \item \textbf{Complexity requirement:} Questions deemed overly simplistic, answerable by a cursory glance at the image without substantive logical engagement, were excluded.
    \item \textbf{Subjectivity and discrepancy check:} If a question is found to be too subjective, or if the validators’ answers significantly differ from the original answer, the question is either revised or removed.
    \item \textbf{Question format diversity:} We ensure a diverse representation of question formats, avoiding the overuse of any particular format of questions.
\end{itemize}

After rigorously applying these quality control measures in several review cycles, our InfiMM-Eval benchmark collection was refined to include 279 high-quality samples.
All samples satisfy our stringent criteria for accuracy, relevance, and cognitive challenge, ensuring a robust and reliable dataset.

\begin{figure}
    \centering
    \includegraphics[width=0.8\textwidth]{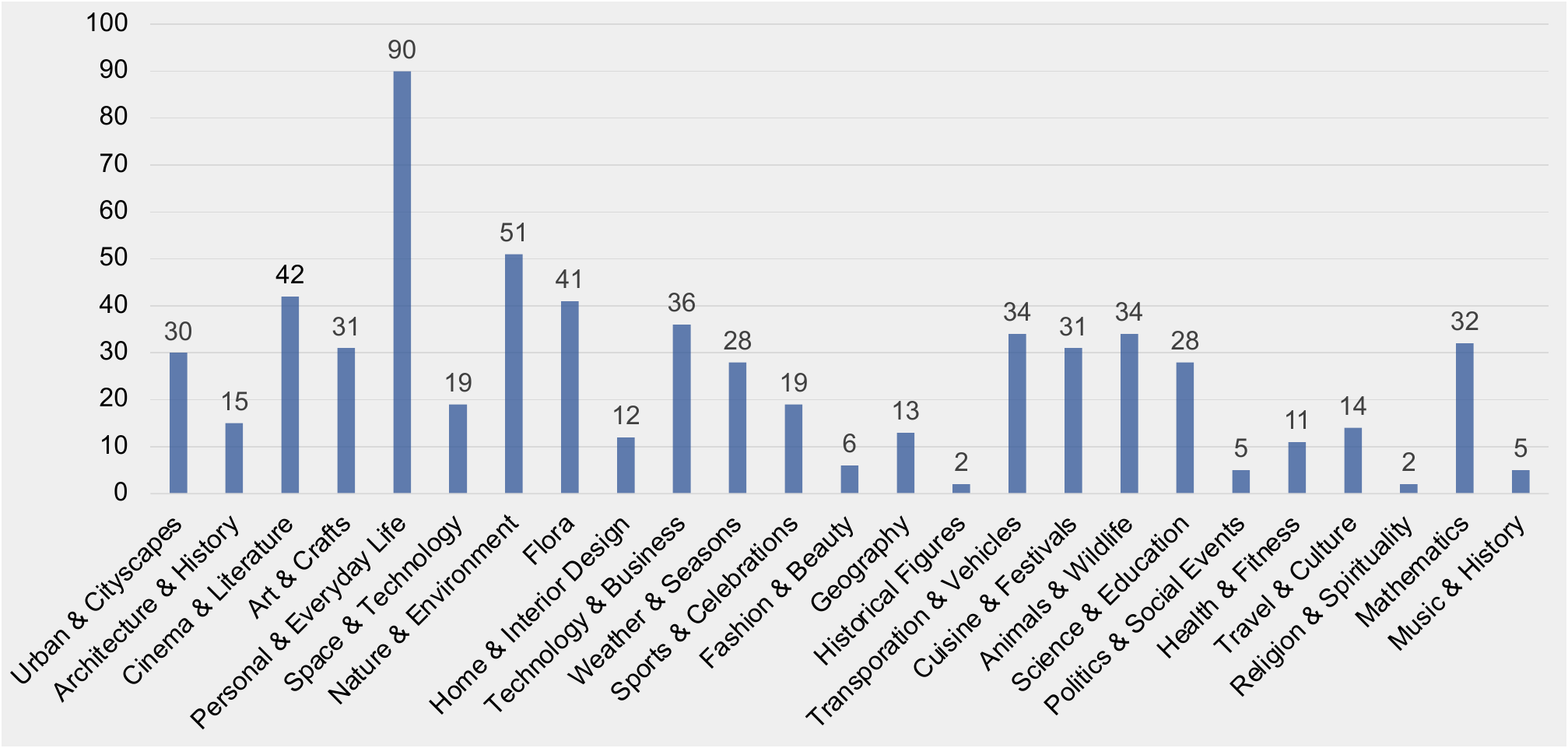}
    \caption{The distribution of visual content categories in InfiMM-Eval benchmark. It is important to highlight that a single image can encompass multiple visual content categories.
    }
    \label{fig:vc_stats}
\end{figure}
\subsection{Dataset statistics}
\label{sec:mm:statistics}
In summary, our InfiMM-Eval benchmark consists of 279 manually curated reasoning questions, associated with a total of 342 images. Out of these, 25 images are adopted from MM-Vet, enriching the diversity and scope of the dataset.

We present a comprehensive statistical analysis of the dataset.
\figureautorefname~\ref{fig:data_cate_stats}~(a) illustrates the distribution across various reasoning types: 49 questions pertain to abductive reasoning, 181 require deductive reasoning, and 49 involve analogical reasoning. 
Furthermore, the dataset is divided into two folds based on reasoning complexity, with 108 classified as ``High'' reasoning complexity and 171 as ``Moderate'' reasoning complexity.
For both abductive and deductive reasoning categories, the ratio of ``High'' to ``Moderate'' questions reasoning complexity is approximately $1:2$, whereas for analogical reasoning, this ratio is closer to $1:1$. 
This distribution underscores the high quality of our benchmark. 
Notably, the dataset includes 23 questions that entail counter-intuitive reasoning (See Appendix for more details), further exemplifying the diversity of our benchmark, as depicted in \figureautorefname~\ref{fig:data_cate_stats}~(b). 
Additionally, as \figureautorefname~\ref{fig:data_cate_stats}~(c) indicates, about $76\%$ (212 out of 279) of the reasoning questions require three or more steps to solve.

\figureautorefname~\ref{fig:vc_stats} demonstrates the diversity of visual content in our image collection, categorized by GPT-4V into a predefined set of concepts. 

\section{Experiments}
In this section, we delineate the experimental settings to assess the reasoning capabilities in contemporary MLLMs. 
Specifically, we furnish a comprehensive description of evaluation baselines and protocols in~\sectionautorefname~\ref{eval_protocol}. 
Subsequent to this, we conduct thorough evaluations and ablation studies on a range of MLLMs using our InfiMM-Eval dataset, as detailed in \sectionautorefname~\ref{sec:results}.

\subsection{Evaluation protocol} 
\label{eval_protocol}
\begin{figure}
    \centering
    \includegraphics[width=0.9\textwidth]{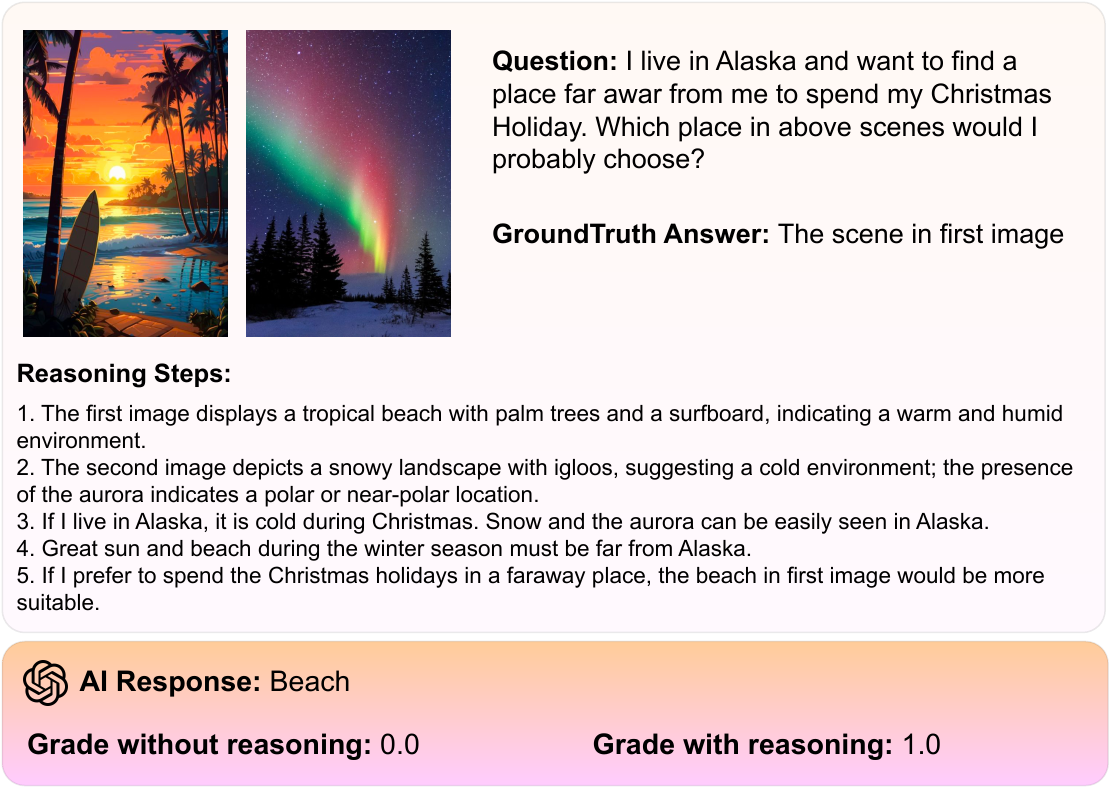}
    \caption{In this example, model can successfully recognize and answer the question, however, due to the nature of open-ended response, the model's response cannot be judged correctly solely based on question and answer. 
    }
    \label{fig:eval_figure}
\end{figure}

Considering the open-ended nature of question-answering in the InfiMM-Eval benchmark and the generative capabilities of modern MLLMs, it becomes clear that solely assessing answer correctness is insufficient, \eg in \figureautorefname~\ref{fig:eval_figure}.
In line with recent studies~\cite{bai2023touchstone,bitton2023visitbench,yu2023mmvet}, we also employ LLMs as evaluators.
However, our approach is distinct in its integration of both questions and answers, as well as the ground-truth and model-predicted reasoning steps into the LLM prompt. 
The inclusion of structured reasoning steps into the LLM context facilitates the accommodation of diverse model outputs and establishes a comprehensive and justified scoring system.
As elaborated in \sectionautorefname~\ref{sec:intro}, our grading protocol awards full marks for direct correctness, with partial scores assigned based on the relevance and logic of reasoning steps.
This method evaluates not only the model's accuracy in answer generation but also offers a an in depth analysis of its decision-making process, illuminating its reasoning pathways.
For any given question $q$, its score $s_q$ falls within the range of $[0, 1]$. 
The overall score $S$ over the entire dataset, which includes considerations of reasoning complexity detailed in \sectionautorefname~\ref{sec:mm:statistics}, is calculated as 
\begin{equation}
    S = \frac{\sum_{x \in M} s_x + 2 \cdot \sum_{y \in H} s_y}{|M| + 2 \cdot |H|} \times 100\%,
\end{equation}
where $M$ and $H$ denote the sets of questions categorized as having ``Moderate'' and ``High'' reasoning complexity.

\begin{table*}[htbp]
    \centering
    \caption{Evaluation results for various MLLMs. Open-source models best performances are indicated with underlines.}
    \resizebox{\textwidth}{!}{
    \setlength{\tabcolsep}{6pt}
    \begin{tabular}{lll|ccc|cc|c}
    \toprule

     \rowcolor{gray!20}  &  & & \multicolumn{3}{c|}{Reasoning Category} &\multicolumn{2}{c|}{Reasoning Complexity}&  \\
     \rowcolor{gray!20}\multirow{-2}{*}{MLLMs}& \multirow{-2}{*}{LLM}& \multirow{-2}{*}{IFT} & Deductive & Abductive & Analogical & Moderate & High & \multirow{-2}{*}{Overall} \\
    \midrule 
         OpenFlamingo-v2~\cite{awadalla2023openflamingo} & MPT-7B~\cite{mpt-7b} & No & 8.88 & 5.3 & 1.11 & 9.47 & 4.72& 6.82 \\
         MiniGPT-v2~\cite{zhu2023minigpt} &LLaMA2-7B~\cite{touvron2023llama} &Yes & 11.02 & 13.28 & 5.69 & 14.45 & 7.27& 10.43 \\
         Fuyu-8B~\cite{fuyu-8b} &Persimmon-8B~\cite{persimmon-8b} &No & 16.42 & 21.49 & 7.78 & 23.06 & 9.91& 15.7 \\
         BLIP-2~\cite{li2023blip2}  &OPT-2.7B~\cite{zhang2022opt} & No & 22.76 & 18.96 & 7.5 & 24.05 & 14.18& 19.31 \\
         InternLM-XComposer-VL~\cite{zhang2023internlmxcomposer} & InternLM-7B ~\cite{2023internlm} & Yes & 26.77 & 35.97 & 18.61 & 39.13 & 17.18& 26.84 \\
         InstructBLIP~\cite{chung2022scaling} &FLAN-T5-XXL~\cite{chung2022scaling}  & Yes& 27.56 & 37.76 & 20.56 & 40.64 & 18.09& 28.02 \\
         LLaMA-Adapter V2~\cite{gao2023llamaadapter} & LLaMA-7B~\cite{touvron2023llama} & No & 28.7 & 46.12 & 22.08 & 41.33 & 21.91& 30.46 \\
         Otter~\cite{li2023otter}  &LLaMA-7B  & Yes & 22.49 & 33.64 & 13.33 & 35.79 & 12.31 & 22.69 \\
         mPLUG-Owl2~\cite{ye2023mplugowl2} & LLaMA-7B & Yes  & 23.43 & 20.6 & 7.64 & 28.79 & 13.18 & 20.05 \\
         IDEFICS-9B-instruct~\cite{laurenccon2023obelisc} &LLaMA-7B & Yes & 22.99 & 34.63 & 20.56 & 34.45 & 16.73& 24.53 \\
         Emu~\cite{sun2023generative} & LLaMA-13B & Yes & 28.9 & 36.57 & 18.19 & 36.18 & 22.0 & 28.24 \\
         LLaVA-1.5~\cite{liu2023llava} & Vicuna-13B~\cite{vicuna2023} & Yes & 30.94 & \textbf{\underline{47.91}} & 24.31 & 47.4 & 21.0& 32.62 \\
         CogVLM-Chat~\cite{wang2023cogvlm} & Vicuna-7B & Yes & 36.75 & 47.88 & 28.75 & \textbf{\underline{55.67}} & 22.5& 37.16 \\
         Qwen-VL-Chat~\cite{bai2023qwenvl} &Qwen-14B \cite{Qwen-VL} & Yes & \textbf{\underline{37.55}} & 44.39 & \textbf{\underline{30.42}} & 46.61 & \textbf{\underline{30.09}} & \textbf{\underline{37.39}} \\
         \midrule
         GPT-4V~\cite{openai2023gpt4} &GPT-4 & Yes & \textbf{74.86} & \textbf{77.88} & \textbf{69.86} & \textbf{93.98} & \textbf{58.98} & \textbf{74.44} \\
    \midrule 
    \bottomrule
    \end{tabular}}
    \label{tab:main_result}
\end{table*}

\subsection{Experimental results and analysis}
\label{sec:results}
Our InfiMM-Eval benchmark evaluates a diverse range of MLLMs, including GPT-4V~\cite{openai2023gpt4}, LLaVA-1.5~\cite{liu2023llava}, Otter~\cite{li2023otter}, MiniGPT-v2~\cite{zhu2023minigpt}, InstructBlip~\cite{dai2023instructblip}, Blip-2~\cite{li2023blip2}, LLaMA-Adapter-V2~\cite{zhang2023llamaadapter}, InternLM-XComposer~\cite{zhang2023internlmxcomposer}, QWen-VL-Chat~\cite{bai2023qwenvl}, Fuyu~\cite{fuyu-8b}, \etc. 
To comprehensively evaluate MLLMs, we apply the Chain-of-Thought (CoT) method~\cite{kojima2022large,wei2022chain}, as well as examine their in-context learning~\cite{li2023otter,alayrac2022flamingo,dong2022survey} capabilities. 
These studies enable us to derive more insightful observations regarding their performance and potential applications.

\subsubsection{Overall results}

The principal findings are encapsulated in \tableautorefname~\ref{tab:main_result}, derived from employing the most effective prompt strategy for each model.
Among all evaluated MLLMs, GPT-4V is particularly noteworthy, exhibiting unparalleled proficiency across all reasoning domains and complexities, with an overall reasoning score of $77.44$. 
In the realm of open-source MLLMs, Qwen-VL-Chat is distinguished as the front-runner with the highest $37.39$ overall score, marginally surpassing CogVLM-Chat.
Additionally, we observe that models fine-tuned with explicit instructions, display superior performance compared to their solely pretrained counterparts, exemplified by models such as Otter and OpenFlamingo-v2.

\tableautorefname~\ref{tab:main_result} further provides a granular breakdown of scores, reflecting the varied reasoning capabilities of the MLLMs. 
GPT-4V continues to exhibit its dominance across all reasoning dimensions. 
Interestingly, most open-source models lag behind GPT-4V, especially in analogical reasoning, which requires not only the detailed comprehension of image content, but also the ability to transfer knowledge from known instances to analogous situations.

\begin{figure*}[htbp]
    \centering
    \includegraphics[width=0.9\textwidth]{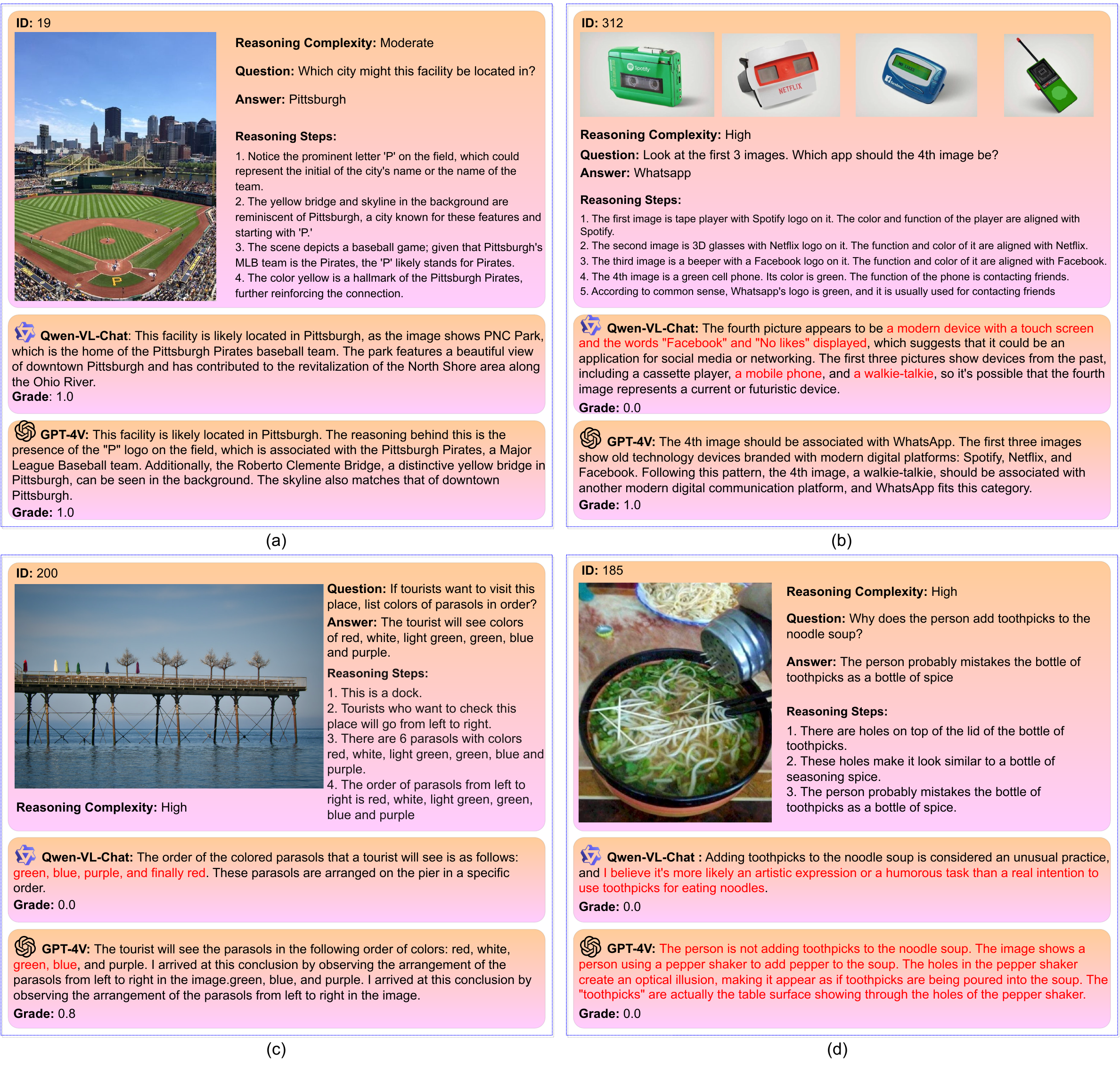}
    \caption{Samples with MLLMs' responses and scores. Hallucinations and errors in model responses are highlighted in \textcolor{red}{red}. }
    \label{fig:easy_hard_samples}
\end{figure*}

To delve deeper, we stratify questions into two levels of complexity: ``Moderate'' and ``High''. 
\figureautorefname~\ref{fig:easy_hard_samples} presents a curated set of examples from our dataset, varying in reasoning complexity, alongside corresponding responses from Qwen-VL-Chat and GPT-4V.
It is noteworthy that GPT-4V consistently outperforms in addressing both moderate and high-complexity questions. Among the open-source models, CogVLM-Chat notably excels in managing moderate complexity questions, whereas Qwen-VL-Chat is particularly adept at handling high-complexity questions.

\subsubsection{Results with chain-of-thought prompt}

In this section, we present a quantitative analysis examining the impact of CoT prompting on MLLMs. 
The results are detailed in \tableautorefname~\ref{tab:cot_prompt_result}.
\begin{table}[htbp]
\begin{minipage}{0.5\textwidth}
    \caption{Comparative evaluation results of MLLMs with and without Chain-of-Thought prompts.}
    \label{tab:cot_prompt_result}
    \resizebox{\textwidth}{!}{
        \begin{tabular}{lccccc}
        \toprule
         \rowcolor{gray!20} MLLMs & CoT & Deductive & Abductive & Analogical & Overall \\
        \midrule 
             \multirow{2}{*}{BLIP-2} & w/o & 22.13 & 18.66 & 5.69 & 18.52 \\
                                    & w & 22.76 & 18.96 & 7.5 & 19.31 \\
             \midrule 
             \multirow{2}{*}{InstructBLIP} & w/o & 25.2 & 34.48 & 16.94 & 25.27 \\
                                        & w & 27.56 & 37.76 & 20.56 & 28.02 \\
             \midrule 
             \multirow{2}{*}{LLaVA-1.5} & w/o & 30.94 & 47.91 & 24.31 & 32.62 \\
                                        & w & 31.18 & 48.51 & 22.78 & 32.6 \\
             \midrule 
             \multirow{2}{*}{Qwen-VL-Chat} & w/o & 38.55 & 45.91 & 22.5 & 36.82 \\
                                        & w & 37.55 & 44.39 & 30.42 & 37.39\\
            \midrule 
             \multirow{2}{*}{GPT-4V} & w/o & 69.88 & 77.88 & 67.08 & 70.72 \\
                                     & w & 74.86 & 77.88 & 69.86 & 74.44\\
        \midrule 
        \bottomrule
        \end{tabular}
    }
\end{minipage}
\hfill
\begin{minipage}{0.5\textwidth}
    \caption{Evaluation results with in-context learning example.}
    \label{tab:icl_prompt_result}
    \resizebox{\textwidth}{!}{
    \begin{tabular}{lccccc}
    \toprule
     \rowcolor{gray!20} MLLMs & ICL & Deductive & Abductive & Analogical & Overall \\
        \midrule 
        \multirow{2}{*}{Otter} & w/o & 22.49 & 33.64 & 13.33 & 22.69 \\
                                    & w & 23.25 & 32.58 & 14.31 & 23.18 \\
        \midrule 
        \multirow{2}{*}{Qwen-VL-Chat 7B} & w/o & 33.73 & 46.82 & 30.28 & 35.32 \\
                                    & w & 38.84 & 44.39 & 27.22 & 37.62 \\
         \midrule 
        \multirow{2}{*}{GPT-4V} & w/o & 74.86 & 77.88 & 69.86 & 74.44 \\
                                 & w & 74.82 & 80.45 & 64.17 & 73.8 \\
    \midrule 
    \bottomrule
    \end{tabular}
    }
\end{minipage}
\end{table}

We adopt a CoT prompting technique similar to that described in \cite{kojima2022large} by appending ``Let's think step by step'' to the end of each question to enhance the reasoning capabilities of the model. 
Our results indicate varied performance changes across different models.
Open-source models generally exhibit a minimal differences in performance, whereas GPT-4V exhibits a notable improvement of $3.7$ with CoT prompts. 
We hypothesize that this phenomenon is attributed to differences in model size and data quality during the instruction-finetuning (IFT) stage of model training. 
The majority of open-source MLLMs are limited by smaller language encoders, typically with less than $14$ billion parameters, inherently constraining their reasoning abilities.
Additionally, the scale and quality of the IFT datasets, commonly used in open-source MLLMs, significantly influence the outcome.
A considerable portion of the IFT data, primarily sourced from VQA~\cite{balanced_vqa_v2}, lacks in reasoning and commonsense knowledge.
This raises an important question about the feasibility of replicating of CoT's success in multimodal contexts.

\subsubsection{Results with in-context learning}

In this section, our focus is on evaluating the in-context learning capabilities of existing MLLMs. 
For this purpose, we have selected three benchmark models for comparison: the high-performing GPT-4V, the leading open-source QWen-VL-Chat, and the Otter. 
It is noteworthy that the Otter distinctively incorporates in-context learning during its training phase.
Specifically, for each query, we randomly select an example from our dataset and integrate it into the prompts during inference. 
This approach is designed to guide and refine the reasoning process of models, ideally enhancing their performance.

As shown in \tableautorefname~\ref{tab:icl_prompt_result}, it is notable that the integration of in-context learning technique does not enhance, and may slightly impair, the performance of the GPT-4V.
In contrast, marginal improvements in performance are observed in the Otter and Qwen-VL-Chat.
These results underscore the complex and diverse nature of the benchmark employed in this study. 
Specifically, for the high-performing GPT-4V, the randomly selected ICL examples might significantly diverge from the test samples.
Conversely, for models with smaller language encoders, such as Otter and Qwen-VL-Chat, which initially demonstrate inferior performance compared to GPT-4V, the inclusion of ICL examples potentially aids in the reasoning process, albeit the impact is relatively limited. 


\subsubsection{Results with LLMs of varied sizes}
\tableautorefname~\ref{tab:llm_impact_result} presents the evaluation results of MLLMs employing LLMs of various sizes.
The dimension of the LLMs is a critical determinant in augmenting the reasoning capabilities of MLLMs.
For instance, considering Qwen-VL\cite{Qwen-VL} as a case study, there is a noticeable increase in the overall reasoning score concurrent with the expansion of the LLM's size. 
Specifically, when the model's size is increased from 7B to 14B parameters, its reasoning score notably increases from $35.32$ to $37.39$.

\begin{table}[ht]
    \centering
    \caption{Evaluation results of models with varied LLM scales.}
    \resizebox{0.8\textwidth}{!}{
    \setlength{\tabcolsep}{6pt}
    \begin{tabular}{lll|cccc}
    \toprule
     \rowcolor{gray!20} Models & LLM & Caption&Deductive & Abductive & Analogical & Overall \\
         \midrule 
        GPT-4 & GPT-4 & - & 5.82 & 5.0 & 2.5 & 5.06 \\
        \hline
        Vicuna-7B &LLaMA-7B& GPT-4V cap. &38.01 & 48.98 & 30.0 & 38.53\\
        Vicuna-13B &LLaMA-13B & GPT-4V cap. &34.42 & 58.78 & 34.69 & 38.75 \\
        SOLAR-0-70b     &LLaMA-70B & GPT-4V cap. &48.56  &64.49  & 33.47 & 48.71 \\
        GPT-4 & GPT-4 & GPT-4V cap.& 54.59 & 66.73 & 45.1 & 55.05 \\
        \hline
        Vicuna-7B(CoT) &LLaMA-7B& GPT-4V cap. &34.42 & 58.78 & 34.69 & 38.75\\
        Vicuna-13B(CoT) &LLaMA-13B &GPT-4V cap.&39.39 & 46.33 & 34.08 & 39.68 \\
        SOLAR-0-70B(CoT)&LLaMA-70B & GPT-4V cap. &54.7  &67.14  & 47.35 & 55.59 \\
        
        GPT-4(CoT) &GPT-4 &LLaVA1.5 cap.& 23.29 & 44.7 & 29.17 & 29.74 \\
        GPT-4(CoT) & GPT-4 & GPT-4V cap.& 55.75 & 66.53 & 51.22 & 56.85 \\
        \midrule 
        \multirow{2}{*}{LLaVa-1.5} & LLaMA2-7B-Chat & - & 27.8 & 33.28 & 21.11 & 27.51 \\
                                    & LLaMA2-13B-Chat & -  & 30.94 & 47.91 & 24.31 & 32.62 \\
        \midrule 
        \multirow{2}{*}{Qwen-VL-Chat} & Qwen-7B & -  & 33.73 & 46.82 & 30.28 & 35.32 \\
                                    & Qwen-14B & -  & 37.55 & 44.39 & 30.42 & 37.39 \\
    \midrule 
    \bottomrule
    \end{tabular}}
    \label{tab:llm_impact_result}
\end{table}

Furthermore, we also report the reasoning capability of standalone language models, such as Vicuna~\cite{vicuna2023} and GPT4~\cite{gpt4systemcard}, by replacing images with their corresponding textual descriptions. 
Prompting GPT-4 directly with only the question resulted in a reasoning score close to 0, as shown in the first row of \tableautorefname~\ref{tab:llm_impact_result}). 
This suggests that the inclusion of visual elements is essential for accurate and effective responses.
As we increase the model size of the LLaMA, from 7B to 70B, there is a noticeable improvement in reasoning scores when utilizing high-quality image descriptions generated by GPT-4V. 
The application of CoT markedly enhances the performance of SOLAR-0-70B, elevating its scores from $48.71$ to $55.59$. 
In contrast, this technique does not produce proportionate enhancements in smaller models, such as those with 7B and 13B.

The GPT-4 model demonstrates optimal reasoning performance when it employs the CoT technique in conjunction with image descriptions generated by GPT-4V. 
A significant reduction in performance is noted when these descriptions are substituted with those produced by LLaVA-1.5. 
Further analysis reveals that the detailed information in GPT-4V's descriptions, including OCR and extensive commonsense knowledge, is crucial for enhancing the ``\textit{multi-modal}'' reasoning capabilities of standalone LLMs.

\section{Conclusion}

In this paper, we introduce InfiMM-Eval, a comprehensive benchmark specifically designed to evaluate complex reasoning capabilities in multi-model language models (MLLMs). Distinct from conventional benchmarks, InfiMM-Eval incorporates not only questions and answers for each data sample but also detailed reasoning steps. For the assessment and grading of open-ended answers and intermediate reasoning procedures, we employ GPT-4. Our evaluation covers a broad spectrum of MLLMs, encompassing both open-source and proprietary models. Additionally, we undertake extensive ablation studies to discern performance disparities among these models. The findings reveal that the current front-runner MLLM, GPT-4V, attains an overall score of 74.44, with a score of 58.98 on more challenging subsets. However, it is noteworthy that the top-performing open-source MLLMs still fall markedly behind GPT-4V in reasoning capabilities. InfiMM-Eval is poised to be a foundational tool for future enhancements in the advanced reasoning capabilities of MLLMs.

\section{Limitations}
In this section, we explore the potential limitations of the existing InfiMM-Eval benchmark. Additionally, we propose avenues for improvement, aiming to enhance its effectiveness and comprehensiveness.

\begin{itemize}
    \item \textbf{Expanding reasoning categories:} The InfiMM-Eval benchmark represents an initial endeavor to scrutinize the capability of deductive, abductive, and analogical reasoning in contemporary MLLMs. 
    Notwithstanding, the spectrum of human reasoning transcends these categories, incorporating more complex forms such as inductive and causal reasoning. 
    Future iterations of this benchmark aim to encompass a broader range of reasoning categories, thereby facilitating a more comprehensive assessment of reasoning capabilities.
    \item \textbf{Enhancing evaluation protocol:} The current InfiMM-Eval benchmark implements a comprehensive evaluation by incorporating intermediate reasoning steps, ultimately producing an overall reasoning score.
    Nevertheless, it is imperative to broaden our evaluation to encompass an in-depth examination of the reasoning process itself. 
    Doing so will yield a deeper insight into the model's reasoning capabilities and render the results more interpretable and accessible to human understanding.
\end{itemize}

\bibliographystyle{unsrt}  
\bibliography{references}  

\clearpage
\appendix
\onecolumn

\vspace*{\stretch{1}}
\begin{center}
    \Large \textbf{Appendix}
\end{center}
\vspace*{\stretch{2}}

\section{Counter-intuitive examples}

\begin{figure*}[hb]
    \centering
    \includegraphics[width=\textwidth]{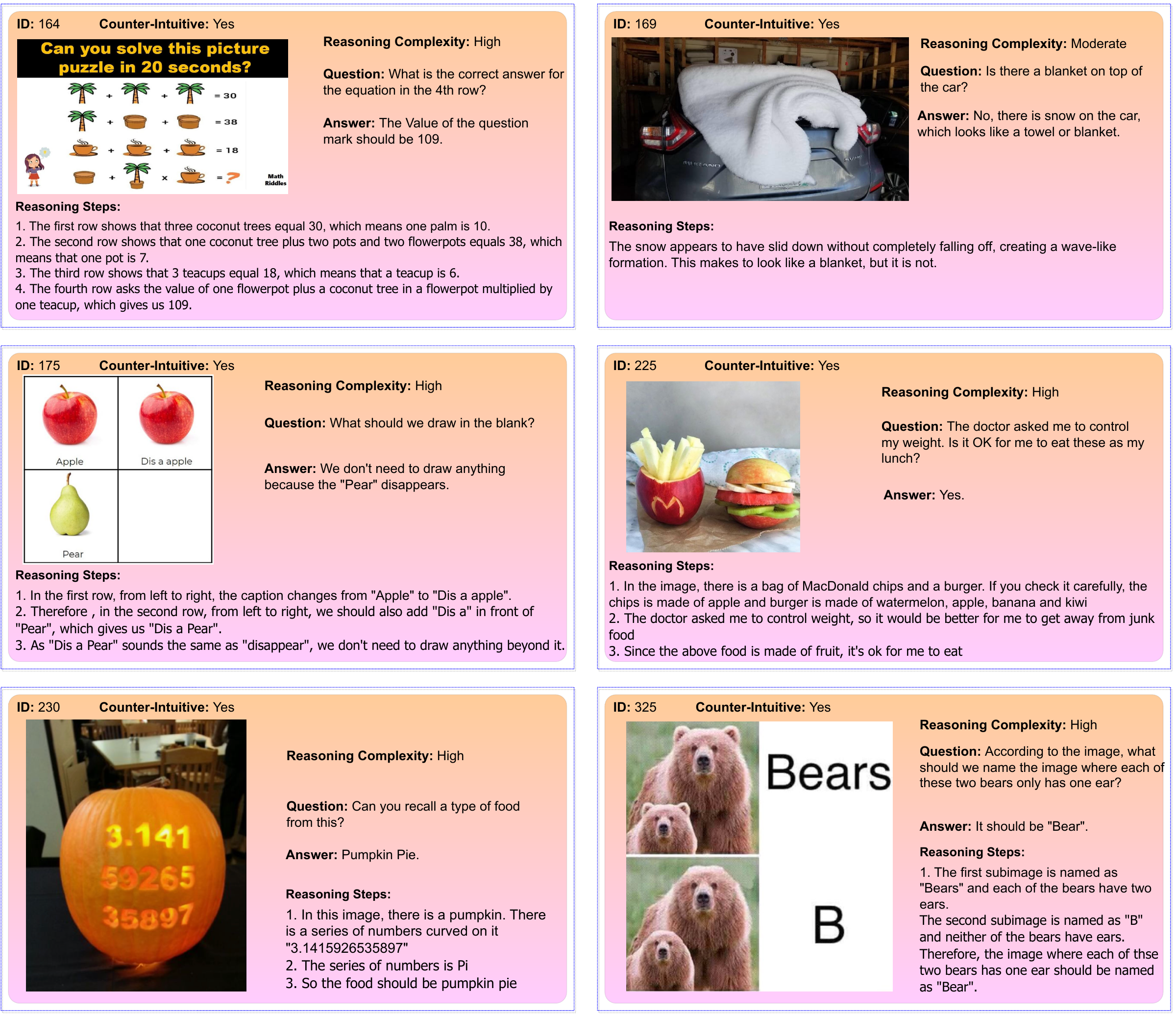}
    \caption{More counter-intuitive examples of InfiMM-Eval.}
    \label{fig:more_example}
\end{figure*}

We provide more counter-intuitive examples of InfiMM-Eval in \figureautorefname~\ref{fig:more_example}.

\section{Model inference prompts}
We list prompts we used for different models in \tableautorefname~\ref{tab:model_prompts}. For Chain-of-thought prompts, we simply add ``Let's think step by step'' at the end of the prompt.

\begin{table*}[htbp]
    \centering
    \caption{Prompts used for evaluations of different models. \{Image\} represents image binary, \{Question\} stands for the questions.}
    \resizebox{\textwidth}{!}{
    \setlength{\tabcolsep}{6pt}
    \begin{tabular}{l|l|l}
    \toprule
    \rowcolor{gray!20} MLLMs & Inference Parameters & Prompts \\
    \midrule 
    GPT-4V & \makecell{temperature: 0.0 \\ top\_p: 0.0 \\ max\_tokens: 256} & \makecell{System Prompt: You are a helpful assistant for helping answer questions. \\ Most questions are related to reasoning. \\
    User Prompt: Here are a list of image detailed descriptions generated by an AI model:\\
    Image 1: \{Image\}\\
    Image 2: \{Image\}\\
    ...\\
    Please answer the following question: \{Question\}} \\
    \midrule
     OpenFlamingo-v2 & \makecell{max\_new\_tokens: 512 \\ num\_beams: 3} & \makecell{\{image\}User: \{question\} GPT:\textless answer\textgreater} \\
     \midrule
     MiniGPT-v2 & \makecell{do\_sample: False \\ max\_new\_tokens: 256} & \makecell{\textless s\textgreater \lbrack INST\rbrack \textless Img\textgreater \{Image\} \textless /Img\textgreater \{Question\} \lbrack /INST\rbrack} \\
     \midrule
     Fuyu-8B & \makecell{max\_new\_tokens:16} &\makecell{  \{Image\}\{Question\}}\\
     \midrule
     BLIP-2  & \makecell{temperature: 1.0 \\ max\_new\_tokens: 20 } &  \makecell{\{Image\} Question:\{Question\} \\ Answer:} \\
     \midrule
     InternLM-XComposer-VL & \makecell{temperature: 1.0 \\ max\_new\_tokens: 1024 } & \makecell{ \textless \textbar User\textbar \textgreater\{Image\} \{Question\}, answer this question \textless eoh\textgreater  \textless \textbar Bot\textbar \textgreater } \\
     \midrule
     InstructBLIP & \makecell{temperature: 1.0 \\ max\_new\_tokens: 128 } & \makecell{  \{Image\}\{Question\}}  \\
     \midrule
     LLaMA-Adapter V2 & \makecell{max\_gen\_len: 256 \\
     temperature: 0.1 \\ top\_k: 0.75} &  \makecell{Below is an instruction that describes a task. \\
        Write a response that appropriately completes the request using a single word or phrase. \\
        Instruction: \{Image\} \{Question\} \\
        Response:}  \\
     \midrule
     Otter  & \makecell{num\_beams:3 \\ max\_new\_tokens:512} & \makecell{\{Image\}User: \{Question\} GPT:} \\
     \midrule
     mPLUG-Owl2 & \makecell{max\_new\_tokens: 256} & \makecell{USER: \{Image\}\{Question\}\\Answer the question using a single word or phrase. ASSISTANT:} \\
     \midrule
     IDEFICS-9B-instruct  & \makecell{temperature: 1.0 \\ max\_new\_tokens:200}  &  \makecell{User: \\ \{image\} \\ \{Question\} \\ Assistant:}\\
     \midrule
     Emu & \makecell{temperature: 1.0 \\ max\_new\_tokens: 128}  &  \makecell{System Prompt: You will be presented with an image: [IMG]\{Image\}[/IMG]. \\ You will be able to see the image after I provide it to you. \\ Please answer my questions based on the given image.\\  \textless \textbar System Prompt\textbar \textgreater USER: \{Question\} ASSISTANT:} \\
     \midrule
     LLaVA-1.5 & \makecell{temperature: 1.0 \\ top\_p: 1.0 \\ max\_tokens: 256} &  \makecell{System Prompt: A chat between a curious user and an artificial intelligence assistant. \\ The assistant gives helpful, detailed, and polite answers to the user's questions. \\ 
     \{Image\}...\{Image\}\\ \{Question\}
     } \\
     \midrule
     CogVLM-Chat &\makecell{temperature: 0.8 \\ max\_new\_tokens: 2048 }   & \makecell{\{Image\}\{Question\}}  \\
     \midrule
     Qwen-VL-Chat & \makecell{do\_sample: False \\ num\_beams: 1 \\ max\_new\_tokens: 100} & \makecell{\textless im\_start\textgreater You are a helpful assistant. \textless im\_end\textgreater \\ Picture 1 \{Image\} \\ Picture 2 \{Image\} \\
     ... \\
     \{Question\} } \\
     \midrule 
    \bottomrule
    \end{tabular}}
    \label{tab:model_prompts}
\end{table*}

\section{Additional ablation study}

In this section, we listed additional ablation studies on InfiMM-Eval.

\subsection{Multi-Images as input results}
Taking multiple images as input is a crucial capability for MLLMs to do multi-round dialogues and interactive step-by-step reasoning. In this section, we explore current MLLMs' multi-image reasoning capability. We compare MLLM's performance by feeding each image seperately and concatenate multiple images horizontally into a single one. Results are listed below in Table~\ref{tab:multi_img_result}.

\begin{table}[h]
    \centering
    \caption{Ablation study results on InfiMM-Eval's subset with multiple images as input. There are 47 samples with multiple images, which contain 27 moderate complexity questions and 20 high complexity questions.}
    \setlength{\tabcolsep}{6pt}
    \begin{tabular}{lcc}
    \toprule
     \rowcolor{gray!20} MLLMs & Concatenate & Score (Multi-Img) \\
    \midrule 
        \multirow{2}{*}{Fuyu-8B} & Yes & 8.21 \\
                                & No & 7.16 \\
        \midrule 
        \multirow{2}{*}{EMU} & Yes & 28.21 \\
                                    & No & 27.76 \\
        \midrule 
        \multirow{2}{*}{GPT-4V} & Yes & 57.61 \\
                                    & No & 71.19 \\
    \midrule 
    \bottomrule
    \end{tabular}
    \label{tab:multi_img_result}
\end{table}

We select Fuyu-8B, EMU and GPT-4V for comparison since these models should support multiple images as input by design. 
Fuyu-8B is a pretrained only model, which does not follow instruction very well, thus cannot achieve good results. 
For EMU, the instruction finetuning data usually do not contain multi-image samples, this could be the reason that there's no evidence of performance improvement. 
For GPT-4V, there is a substantial drop after concatenating images together. 
If the trained model internally cuts the image into patches for processing, such as Fuyu-8B, concatenating images into a single image might impact their input patches and lead to worse performance. 

\end{document}